\documentclass{article}

\PassOptionsToPackage{numbers,compress}{natbib}

\usepackage[preprint]{neurips_2026}

\usepackage[utf8]{inputenc} 
\usepackage[T1]{fontenc}    
\usepackage{hyperref}       
\usepackage{url}            
\usepackage{booktabs}       
\usepackage{amsfonts}       
\usepackage{nicefrac}       
\usepackage{microtype}      
\usepackage{xcolor}         
\usepackage{tabularx}
\usepackage{graphicx}
\usepackage{amsmath}
\usepackage{multirow}

\title{VSRo-200: A Romanian Visual Speech Recognition Dataset for Studying Supervision and Multimodal Robustness}

\author{%
  Iulia-Maria ~Udrea, Alexandra Diaconu, Bogdan Alexe\\
  Department of Computer Science,
  University of Bucharest\\
}

\begin{document}

\maketitle

\begin{abstract}
We introduce VSRo-200, the first large-scale dataset for visual speech recognition (lip reading) in Romanian, comprising 200 hours of real-world podcast videos. All samples are annotated with pseudo-labels generated by a fine-tuned Romanian ASR model, while a subset of 100 hours is additionally transcribed by humans, enabling controlled analysis of supervision quality under a unified framework.
Building on this dataset, we establish a benchmark for visual speech recognition in low-resource settings. We systematically study the impact of supervision quality, showing that while human annotations provide better performance at fixed data scales, pseudo-labels enable continued improvements through scalability. We further evaluate robustness under domain shift using curated out-of-distribution (OOD) test sets, and analyze audio-visual speech recognition (AVSR) under noisy conditions, where multimodal fusion significantly improves robustness compared to audio-only models.
Finally, we demonstrate that representations learned on VSRo-200 transfer effectively to the LRRo benchmark for isolated word recognition, substantially outperforming previously reported results.
Overall, VSRo-200 provides a new testbed for studying supervision, domain generalization, and multimodal fusion in low-resource visual speech recognition.
\end{abstract}

\section{Introduction}
Visual speech recognition (VSR), or lip reading, aims to transcribe speech solely from visual input. The task is inherently challenging~\cite{assael2016lipnet} due to the ambiguity of visual speech, where multiple phonemes map to the same viseme, making certain words visually indistinguishable (e.g., “ma”, “pa”, “ba”).

In recent years, substantial progress has been achieved for English VSR, driven by the availability of large-scale datasets and advances in deep spatio-temporal models. Early works focused on word-level recognition using datasets such as LRW~\cite{chung2016lip}, while later efforts extended to sentence-level transcription using larger corpora such as LRS2~\cite{chung2017lip} and LRS3~\cite{afouras2018lrs3}. These developments, combined with modern architectures ~\cite{petridis2018end, martinez2020lipreading, ma2022training}, have enabled strong performance in high-resource settings.

Beyond English, several works have explored cross-lingual transfer, showing that visual speech representations learned on high-resource languages can generalize to other languages such as German~\cite{schwiebert2022multimodal}. Recently, multilingual approaches have demonstrated that training on multiple languages jointly can significantly improve performance~\cite{ma2022visual}. In particular, large-scale web video datasets with automatic labeling~\cite{prajwal2025scaling, yeo2024visual} enable training on thousands of hours across multiple languages such as Spanish, French, Italian, and Portuguese, demonstrating the effectiveness of weak supervision at scale.

Despite this progress, low-resource languages remain largely underexplored. In the case of Romanian, existing work is limited to small-scale datasets and word-level recognition tasks~\cite{jitaru2020lrro}, as well as transfer learning approaches that leverage high-resource languages~\cite{jitaru2021toward, manescu2023end}, which do not reflect realistic continuous speech scenarios. Furthermore, current approaches either rely on cross-lingual transfer or fully automatic labeling pipelines, highlighting the lack of large-scale, high-quality annotated data for Romanian VSR.

More broadly, existing VSR datasets exhibit a clear trade-off between scale and annotation quality: they are either relatively small but carefully annotated~\cite{assael2016lipnet} or large-scale collections relying predominantly on pseudo-labels generated by automatic speech recognition (ASR) systems~\cite{prajwal2025scaling, yeo2024visual}. The success of recent ASR models such as Whisper~\cite{radford23apmlr} has further accelerated this trend by enabling large-scale automatic annotation.

While recent large-scale efforts demonstrate that pseudo-labels can enable effective scaling, they do not provide a controlled setting to analyze the impact of supervision quality. As a result, a fundamental question remains insufficiently explored: \emph{what is the role of supervision quality in visual speech recognition, particularly in low-resource settings?}

In this work, we introduce \textit{VSRo-200}, the first large-scale dataset for Romanian VSR, comprising 200 hours of real-world podcast videos. All data is annotated with pseudo-labels generated by a fine-tuned Romanian ASR model, while a subset of 100 hours is additionally transcribed by humans, enabling controlled comparison between weak and full supervision.
Using this dataset, we study the impact of supervision quality, robustness to domain shift through OOD test sets, and audio-visual fusion under noisy conditions. 

Our experiments yield three key findings. First, pseudo-labels enable effective scaling, approaching the performance of human annotations at larger data sizes. Second, performance under domain shift remains challenging, with degradation driven by factors such as vocabulary mismatch and visual variability. Third, audio-visual fusion improves performance in noisy environments, highlighting the complementary nature of visual and acoustic cues.

\section{Related Work}

\noindent{\bf Datasets for VSR.}
Progress in VSR has been strongly driven by the availability of benchmark datasets (see Table~\ref{tab:vsr_datasets}). Early sentence-level work relied on controlled datasets such as GRID~\cite{cooke2006audio}, where utterances follow a fixed grammar and are recorded in laboratory conditions. Large-scale English resources later shifted toward more realistic settings: LRW~\cite{chung2016lip} introduced word-level recognition from broadcast television, while LRS2~\cite{chung2017lip} and LRS3~\cite{afouras2018lrs3} enabled sentence-level transcription from real-world video sources. These datasets have become standard benchmarks for modern VSR systems, but they are predominantly English and assume access to large-scale annotated data.

\noindent{\bf Beyond English and multilingual VSR.}
Several works have extended VSR beyond English. LRW-1000~\cite{yang2019lrw} introduced a large-scale Mandarin word-level benchmark, while GLips\cite{schwiebert2022multimodal} provides a German word-level dataset and studies transfer learning from other languages. More recent multilingual approaches~\cite{ma2022visual} study joint training across languages and show that multilingual data can improve VSR performance. Large-scale pseudo-labeling pipelines~\cite{prajwal2025scaling,yeo2024visual} further improve scalability by using ASR systems such as Whisper~\cite{radford23apmlr} to automatically annotate web videos. In particular, MultiVSR~\cite{prajwal2025scaling} scales multilingual VSR to approximately 12,000 hours of automatically labeled data across 13 languages. These works demonstrate the effectiveness of weak supervision at scale, but they do not explicitly isolate the effect of annotation quality.

\noindent{\bf Romanian lip reading.}
Romanian VSR remains under-resourced. LRRo~\cite{jitaru2020lrro} is the first Romanian lip-reading dataset, containing $\sim$26 hours of data collected in both wild and laboratory settings. However, LRRo is restricted to isolated word recognition with a small vocabulary. Subsequent work~\cite{jitaru2021toward, manescu2023end} explores transfer learning and cross-lingual adaptation to compensate for the lack of Romanian data. These approaches confirm that Romanian VSR suffers from data scarcity, but they remain limited to word-level classification and do not support continuous sentence-level transcription.

\noindent{\bf AVSR and robustness.} Beyond visual-only models, several works~\cite{petridis2018end, ma2023auto} have explored AVSR, showing that combining visual and acoustic modalities can improve robustness, particularly in noisy environments. However, these approaches are typically evaluated in high-resource settings and do not explicitly address low-resource scenarios. Robustness to domain shift is another challenge in VSR, as models trained on curated datasets often degrade when applied to real-world data. While large-scale datasets partially mitigate this issue, systematic evaluation under OOD conditions remains limited.

\noindent{\bf Positioning of VSRo-200.}
As summarized in Table~\ref{tab:vsr_datasets}, existing resources reveal a clear gap. English VSR benefits from large sentence-level datasets, multilingual VSR increasingly relies on large pseudo-labeled corpora, and Romanian VSR is limited to small word-level datasets. VSRo-200 addresses this gap by introducing a 200-hour Romanian sentence-level dataset with both human and pseudo-labeled annotations. Unlike purely pseudo-labeled multilingual datasets, VSRo-200 is designed to study supervision quality, enabling controlled comparison between models trained with human annotations and models trained with pseudo-labels, as well as robustness evaluation under OOD conditions.

\begin{table*}[t]
\centering
\setlength{\tabcolsep}{2.5pt}
\caption{Comparison of representative VSR datasets. 
We report language, recognition unit, data source, dataset scale, vocabulary size, and supervision type (M = manual, P = pseudo, U = unlabeled). VSRo-200 is the first large-scale Romanian sentence-level dataset with both human and pseudo-label annotations, enabling controlled study of supervision quality.}
\label{tab:vsr_datasets}
\begin{tabular}{lcccccc}
\toprule
\textbf{Dataset} & \textbf{Language} & \textbf{Unit} & \textbf{Source} & \textbf{Hours} & \textbf{Vocabulary} & \textbf{Supervision} \\
\midrule
\midrule
GRID~\cite{cooke2006audio} & EN & Sentences & Lab & 27.5 & 51 & M \\
LRW~\cite{chung2016lip} & EN & Words & TV & $\sim$165 & 500 & M \\
LRS2~\cite{chung2017lip} & EN & Sentences & TV & $\sim$224 & 40k+ & M \\
LRS3~\cite{afouras2018lrs3} & EN & Sentences & TED/YT & $\sim$400 & 50k+ & M \\
\midrule
\midrule
LRW-1000~\cite{yang2019lrw} & ZH & Words & TV & $\sim$57 & 1k & M \\
GLips~\cite{schwiebert2022multimodal} & DE & Words & Parl. & -- & 500 & M \\
\midrule
\midrule
mTEDx~\cite{salesky21_interspeech} & Multi & Sentences & TEDx & $\sim$285 & Large & M \\
VoxCeleb2~\cite{chung18b_interspeech} & Multi & Speech Segments & YT & 2440 & -- & U \\
AVSpeech~\cite{ephrat2018looking} & Multi & Speech Segments & YT & 4700 & -- & U \\
MultiVSR~\cite{prajwal2025scaling} & Multi & Sentences & YT & $\sim$12000 & Large & P \\
\midrule
\midrule
LRRo~\cite{jitaru2020lrro} & RO & Words & Mixed & $\sim$26 & 21--48 & M \\
\textbf{VSRo-200 (ours)} & \textbf{RO} & \textbf{Sentences} & \textbf{Podcasts (YT)} & \textbf{200} & \textbf{Large} & \textbf{M+P} \\
\bottomrule
\end{tabular}
\end{table*}

\section{Data Collection and Preprocessing}
\label{sec:data}

We construct VSRo-200 from Romanian podcast videos collected from YouTube. Each podcast episode typically features a host and a single guest in a fixed multi-camera setup, alternating between views of the host, the guest, and wide shots. To maximize identity diversity, we select episodes with a single guest and ensure that each individual appears in only one episode. The resulting dataset contains 235 speakers in total (20 reserved for testing), comprising 64,710 video clips, with 35,102 clips from female speakers and 29,608 from male speakers.

Our goal is to segment these long-form recordings (typically $\sim$2 hours) into short, sentence-level clips of 3–30 seconds. We first partition each video into shots based on camera angle changes using \texttt{PySceneDetect}~\cite{pyscenedetect}, resulting in segments with consistent visual framing.

\noindent{\bf Visual filtering.}
We filter the extracted shots to retain only segments containing a single visible speaker. Using the \texttt{InsightFace} model~\cite{deng2019arcface, guo2021sample}, we detect faces and compute feature embeddings for each shot (based on its midpoint frame). Shots containing multiple faces are discarded. To further remove host appearances, we compare detected face embeddings against a precomputed bank of host embeddings using cosine similarity, and discard shots exceeding a threshold of 0.5. This yields candidate segments containing only the guest.

\noindent{\bf Audio filtering and segmentation.}
Although the retained shots contain only the guest visually, speech may still originate from the host off-camera. To address this, we apply speaker diarization using \texttt{Pyannote}~\cite{bredin2020pyannote} to obtain per-speaker timestamps. We identify the guest as the speaker with the largest cumulative speaking duration within each shot, and use diarization boundaries together with detected pauses to segment the audio into clips of 3–30 seconds containing only guest speech.

\noindent{\bf Face tracking and synchronization.}
The resulting segments are further processed using a pipeline inspired by MultiVSR~\cite{prajwal2025scaling}. We apply \texttt{S3FD}~\cite{zhang2017s3fd} face detection and tracking, discard clips with face resolution below 96$\times$96 pixels, and use \texttt{SyncNet}~\cite{chung2016out} to verify audio-visual synchronization. Clips with a synchronization offset larger than $\pm$10 frames are removed. The final outputs are face crops of size 224$\times$224 at 25 FPS with aligned 16 kHz audio.

\noindent{\bf Pseudo-label generation.}
The source podcast videos do not provide transcripts. To obtain supervision at scale, we generate pseudo-labels using a Whisper-large model~\cite{radford23apmlr} fine-tuned for Romanian~\cite{diaconu2026ron3ws}. These automatically generated transcriptions are associated with all extracted clips and serve as supervision for downstream training.
This pseudo-labeling strategy enables full coverage of the dataset, while allowing controlled comparison with human annotations on a subset of the data.


\subsection{Human Annotation}

In addition to pseudo-labels, we annotate a subset of 100 hours with high-quality human transcriptions. This subset is selected from visually clean segments, ensuring clear visibility of the speaker’s mouth and the absence of overlapping speakers.

\noindent{\bf Annotation protocol.}
Human annotators are instructed to produce verbatim transcriptions of the spoken content, following standard orthographic conventions for Romanian. The guidelines emphasize consistency in punctuation, normalization of numbers and abbreviations, and preservation of spoken structure, while avoiding unnecessary corrections or reinterpretations.

\noindent{\bf Alignment with pseudo-labels.}
The human-annotated subset overlaps with the pseudo-labeled data, enabling direct comparison between annotation types while keeping the underlying data distribution consistent. This design allows us to study the impact of annotation quality on model performance.

Ethical considerations and dataset release details are provided in the supplementary material.

\section{Experimental Evaluation}
\label{sec:exp-eval}

We evaluate VSRo-200 through a series of experiments analyzing supervision quality and robustness across multiple dimensions, including noise, domain shift, and generalization across datasets.

\subsection{Task and Evaluation Protocol}

We consider sentence-level VSR, where the goal is to transcribe speech from video frames, and report performance using word error rate (WER). 
We evaluate models under different supervision regimes, including training with human annotations and pseudo-labels. We further assess robustness through: (i) noise conditions via audio-visual modeling; (ii) domain shift using OOD data; (iii) cross-dataset generalization on an external Romanian benchmark.


\subsection{Experimental Setup}

\noindent{\bf Data.}
The human-annotated subset (100 hours) of VSRo-200 is used to train models under clean supervision, while the full dataset (200 hours) with pseudo-labels is used for weak supervision.

\noindent{\bf Models.}
We adopt a transformer-based VSR model following~\cite{prajwal2025scaling}. Visual features are first extracted using a pre-trained Visual Transformer Pooling (VTP) network, which encodes short-term spatio-temporal information from the input video. These features are then processed by a Transformer encoder-decoder architecture that predicts the output text sequence in an auto-regressive manner.
For AVSR, we extend the model by incorporating audio features extracted from the speech signal.

\noindent{\bf Training details.}
All models are trained on face crops extracted from video sequences sampled at 25 FPS. We follow~\cite{prajwal2025scaling}, using a sequence-to-sequence Transformer encoder-decoder architecture with a shared tokenizer and the same hyperparameters.
The main difference is that MultiVSR trains on fixed-length segments (approximately 14 seconds) obtained using timestamp alignment, while we train directly on full clips, as our data does not include precise timestamps. In practice, we observe no significant difference in performance between the two strategies.
%
Optimization is performed using Adam with a learning rate of $10^{-4}$, reduced on plateau, and models are trained until convergence. For audio-visual experiments, audio is sampled at 16 kHz and processed jointly with visual features.
Training settings are kept consistent across experiments to ensure fair comparison between supervision regimes and evaluation conditions. All experiments were run on a system with an NVIDIA H100 GPU (80GB), Intel Xeon Platinum 8480+ CPU, and 2TB RAM.

\subsection{Impact of Supervision Quality}

We first analyze the effect of supervision quality by comparing models trained on human annotations with those trained on pseudo-labels.

\begin{figure}[t]
\centering
\includegraphics[width=\linewidth]{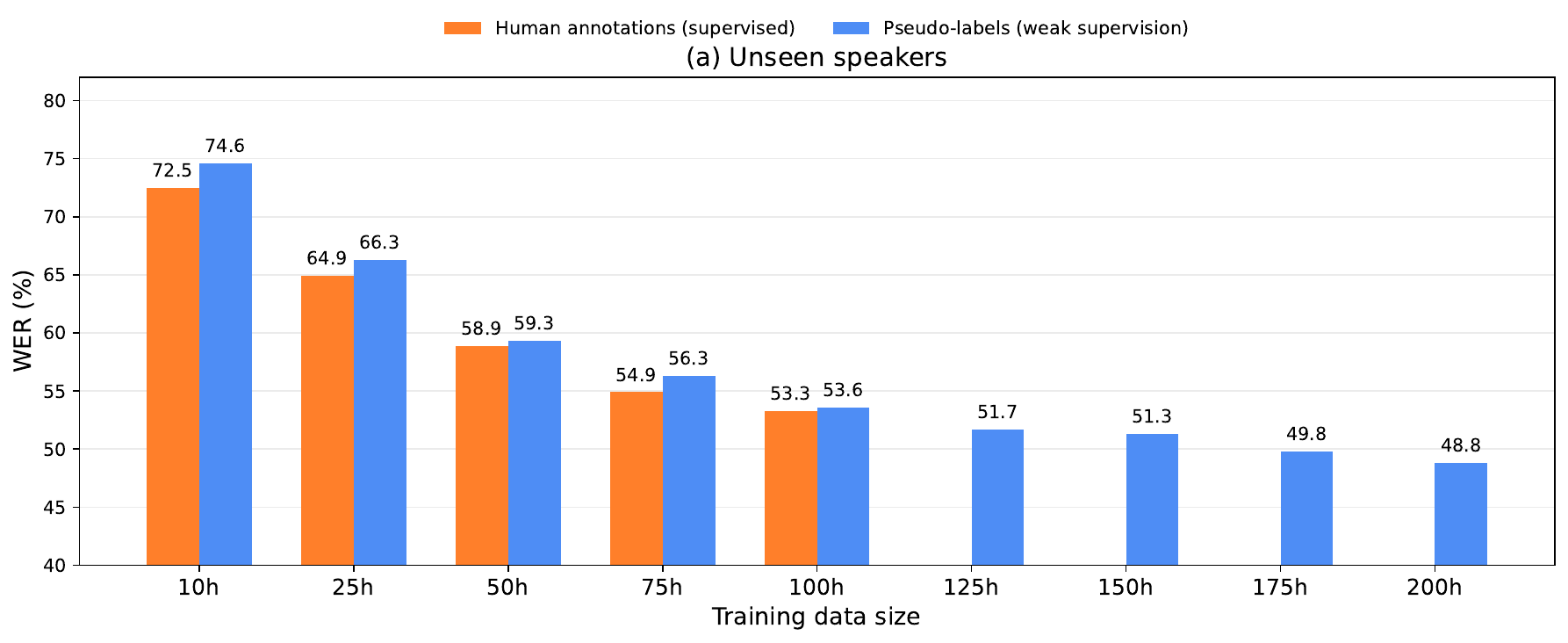} \\
\includegraphics[width=\linewidth]{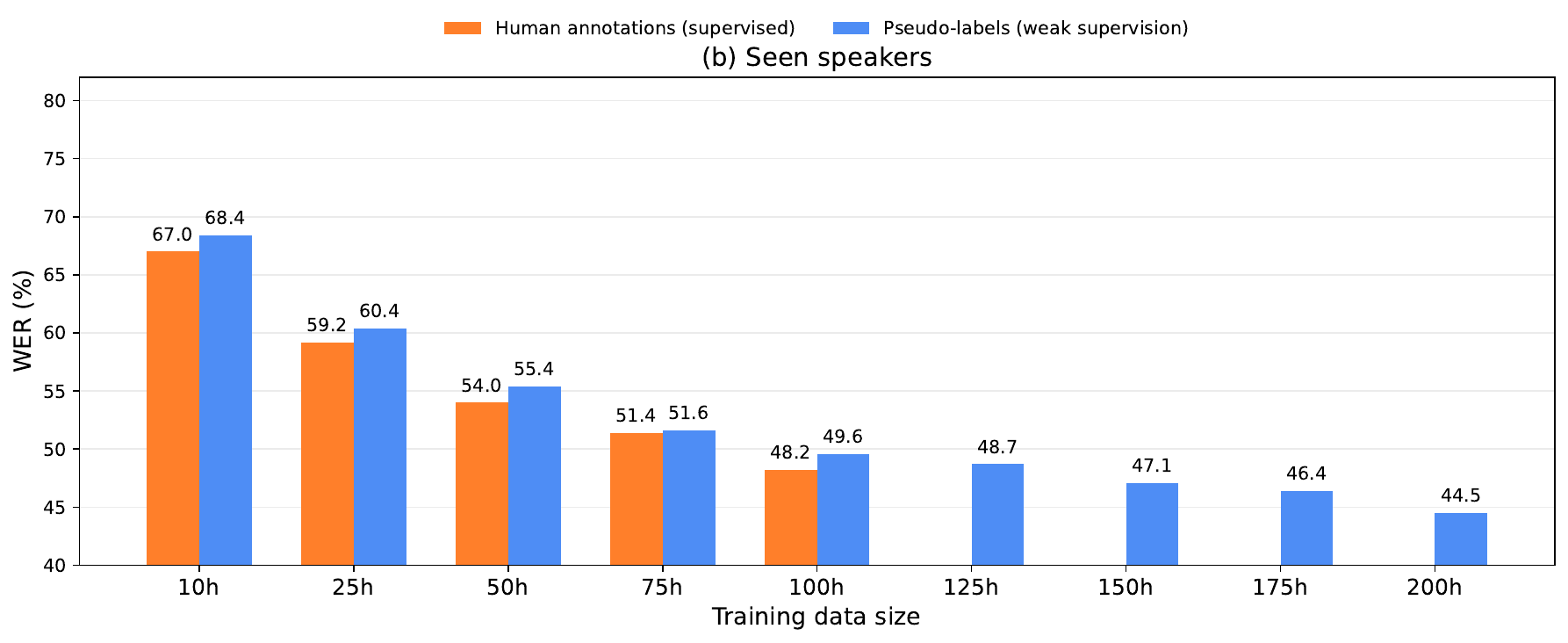}

\caption{Impact of supervision quality and training data scale on VSR performance under different evaluation settings: (a) unseen speakers and (b) seen speakers.} 
\label{fig:supervision_scaling_combined}
\end{figure}

\noindent{\bf Setup.}
We train models using 10, 25, 50, 75, and 100 hours of data for both types of supervision. For pseudo-labels, we additionally train models using 125, 150, 175, and 200 hours, since pseudo-labels are available for the full dataset. 
%
%
Evaluation is performed on two fixed test sets, each of $\sim$1 hour of data, for a total of 2 hours. The first test set (\textit{unseen speakers}) contains 20 speakers (10 male and 10 female) that are disjoint from the training data. The second test set (\textit{seen speakers}) also contains 20 speakers (10 male and 10 female), but these identities are present in the training data for all models. Both test sets are balanced and kept identical across all experiments to ensure fair comparison.

%

\noindent{\bf Inference.}
During decoding, we employ beam search with an n-gram repetition constraint to reduce degenerate outputs. Specifically, the decoder prevents repeated token sequences of length $N=5$ by masking candidate tokens that would complete a previously generated n-gram.

\noindent{\bf Results and discussion.}
Figure~\ref{fig:supervision_scaling_combined} shows that increasing the training data consistently improves performance for both supervision types under both evaluation settings. As expected, models achieve lower WER when evaluated on seen speakers compared to unseen speakers, reflecting the challenge of speaker generalization in VSR.
At comparable data scales, models trained with human annotations achieve slightly lower WER than those trained with pseudo-labels (e.g., 53.3\% vs.\ 53.6\% at 100 hours on unseen speakers). Importantly, the performance gap between the two supervision types decreases as the amount of training data increases. While human annotations provide a clear advantage at smaller scales, this advantage becomes marginal at larger scales, indicating that increased data volume can compensate for the noise in pseudo-labels.
To assess statistical stability, we repeat the 100-hour experiment three times with different data shuffles. Results show low variance for both supervision types (e.g., $53.21 \pm 0.30$ vs.\ $53.82 \pm 0.14$ WER on unseen speakers and $48.29 \pm 0.18$ vs. $49.66 \pm 0.05$ WER on seen speakers for human and pseudo-labels, respectively), confirming that the observed differences are consistent and not due to random variation.
Moreover, pseudo-labels enable training beyond the human-annotated subset. Scaling from 100 to 200 hours of pseudo-labeled data further reduces WER (e.g., from 53.6\% to 48.8\% on unseen speakers), demonstrating the benefits of large-scale weak supervision. Notably, these trends are consistent across both seen and unseen evaluation settings.
Overall, these results highlight a trade-off between annotation quality and data scale: human annotations provide stronger supervision at matched sizes, while pseudo-labels enable continued gains through scalability. At the same time, the performance gap between seen and unseen speakers underscores the importance of evaluating generalization, validating VSRo-200 as a resource for studying both supervision quality and robustness in low-resource VSR.

\subsection{Robustness to Acoustic Noise}
\label{avsr}

\noindent{\bf Setup.}
We evaluate AVSR under controlled noisy conditions by corrupting the audio signal with two types of noise: Gaussian noise and babble noise~\cite{snyder2015musan}. 
For each noise type, we consider signal-to-noise ratio (SNR) levels of $-5$, $0$, $5$, $10$, and $15$ dB.
We compare three systems: (i) audio-only ASR using Whisper (in both zero-shot and fine-tuned variants); (ii) visual-only VSR using MultiVSR trained on the full 200-hour pseudo-labeled dataset (corresponding to our best-performing visual model); (iii) audio-visual fusion obtained by combining the two modalities.
Evaluation is performed using WER on a subset of 100 clips ($\approx$16 minutes) sampled from the unseen-speaker test set.

\vspace{0.3em}
\noindent{\bf Audio-visual fusion.}
We combine audio and visual predictions using a token-level probabilistic fusion strategy.
At each decoding step $t$, both the audio model (Whisper) and the visual model (MultiVSR) produce a probability distribution over a shared vocabulary
$P_{\text{audio}}(w_t)$ and $P_{\text{video}}(w_t)$. Since both models use the same tokenizer, their outputs are directly comparable and can be fused. A fixed interpolation between the two modalities would fail to account for their varying reliability. 
Instead, we adopt a {\it confidence-aware dynamic fusion} strategy, where the contribution of each modality is determined by its prediction uncertainty. We quantify uncertainty using the entropy of each distribution:
$H(P) = - \sum_{w} P(w)\log P(w).$
Low entropy indicates confident predictions, while high entropy reflects uncertainty. 
We define a dynamic weight $\alpha_t$ as:
\begin{equation}
\alpha_t = \frac{H(P_{\text{video}})}{H(P_{\text{audio}}) + H(P_{\text{video}})}
\label{eq:alpha}
\end{equation}
which leads to the fused distribution:
\begin{equation}
P_{\text{fused}}(w_t) = \alpha_t P_{\text{audio}}(w_t) + (1 - \alpha_t) P_{\text{video}}(w_t)
\label{eq:fusion}
\end{equation}

This formulation assigns higher weight to the modality with lower entropy (i.e., higher confidence). 
For example, when the audio signal is degraded by noise, $H(P_{\text{audio}})$ increases, reducing $\alpha_t$ and shifting more weight toward the visual modality. At each decoding step, the predicted token is obtained as $w_t = \arg\max_w P_{\text{fused}}(w_t)$ and is fed back as context to both models for the next step, corresponding to greedy decoding. 
In practice, we extend this formulation to beam search by applying the same fusion at each step while maintaining the top-$K$ hypotheses based on cumulative log-probability. The final transcription is selected as the highest-scoring complete sequence. All results reported in this work use beam search with beam size $K=5$. An overview of the fusion mechanism is provided in the supplementary material.


\begin{figure*}[t]
\centering

\includegraphics[width=0.49\linewidth]{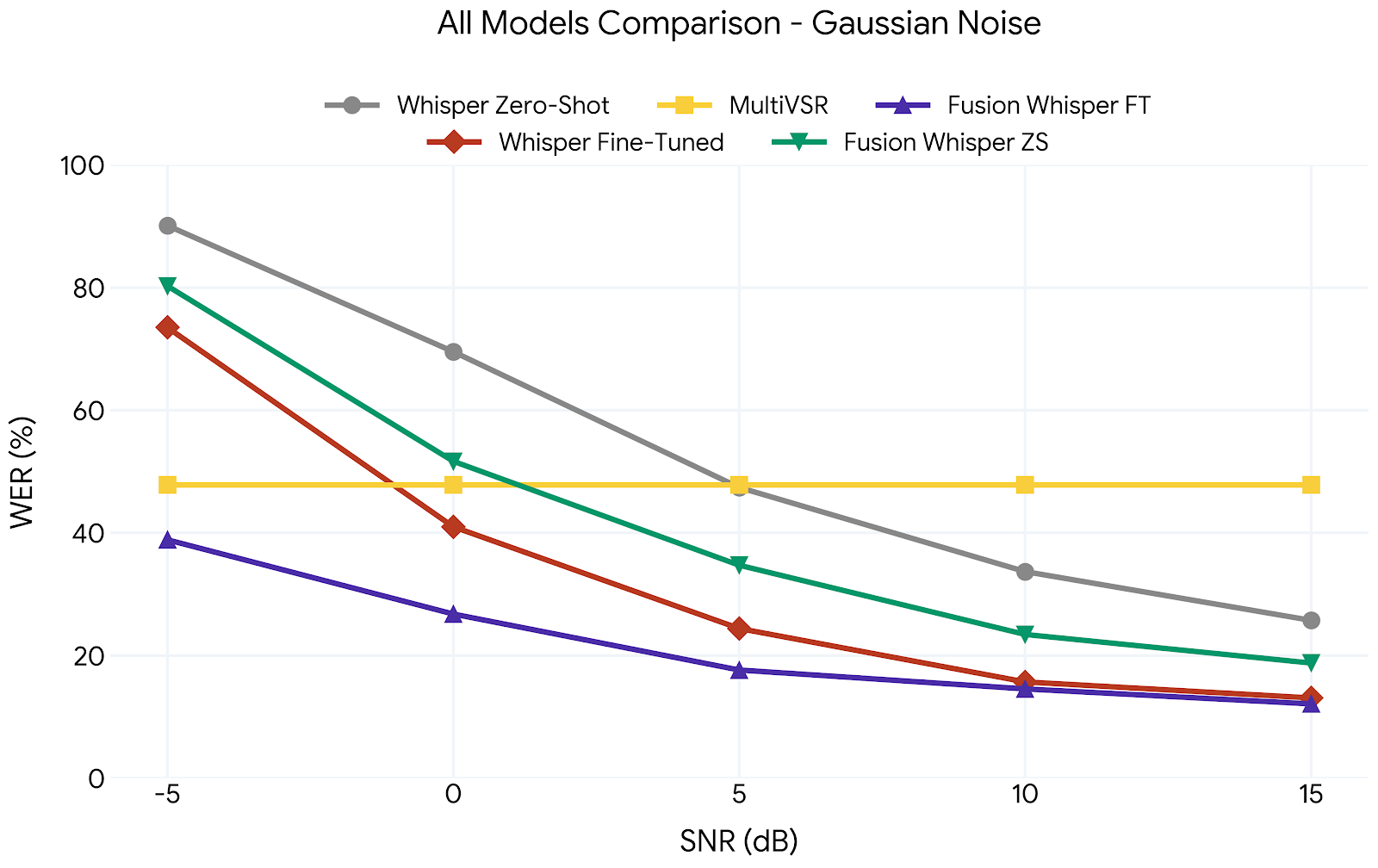}
\includegraphics[width=0.49\linewidth]{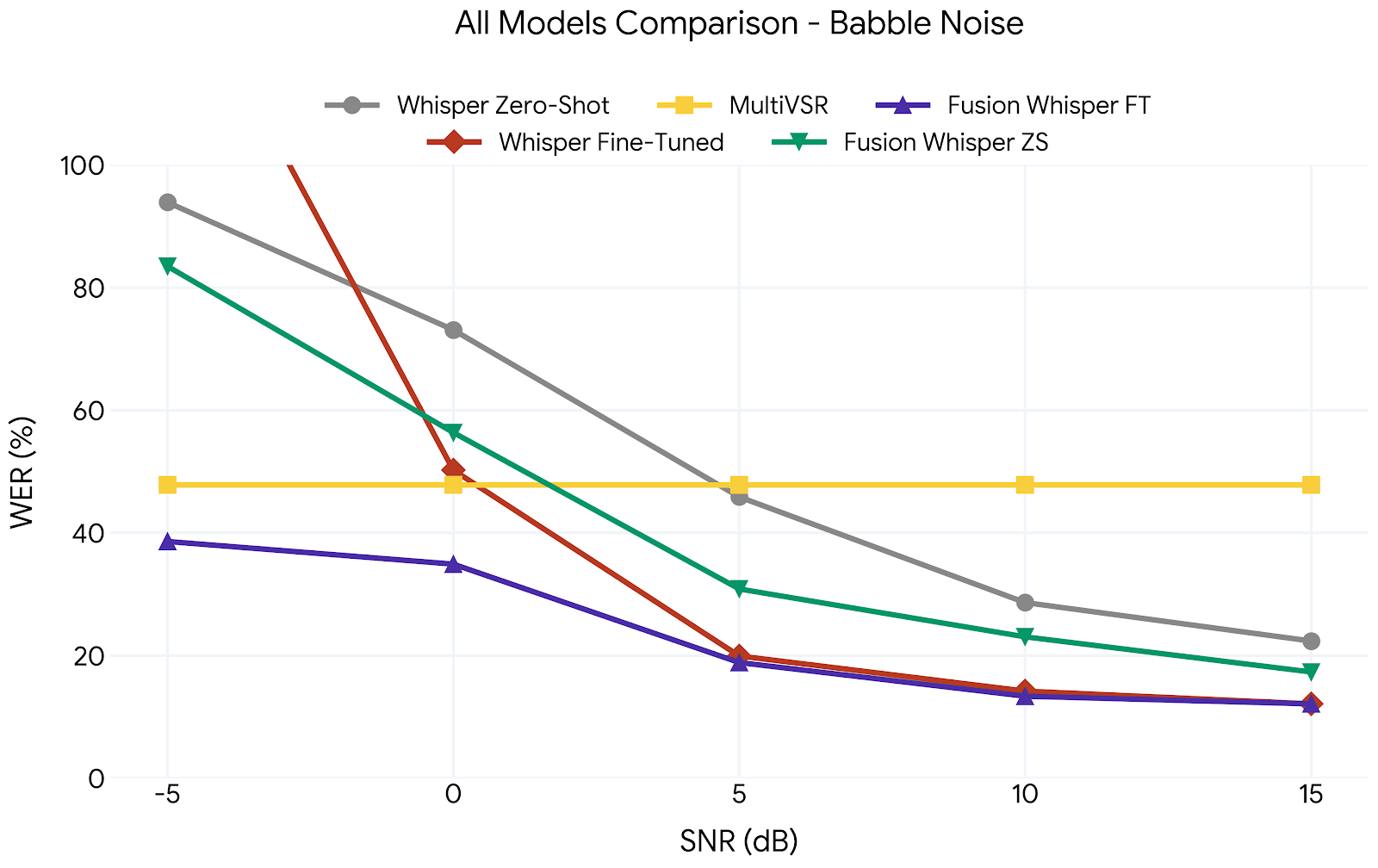}

\caption{Robustness to acoustic noise under different conditions.
(a) Gaussian noise with zero-shot Whisper and fine-tuned Whisper;
(b) babble noise with zero-shot Whisper and fine-tuned Whisper.}
\label{fig:avsr_noise}
\end{figure*}

\vspace{0.3em}
\noindent{\bf Results.}
Figure~\ref{fig:avsr_noise} summarizes performance across noise types and SNR levels.

We observe that audio-only performance degrades rapidly as noise increases. Under Gaussian noise at $-5$ dB, Whisper zero-shot reaches 90.11\% WER, while the fine-tuned model reaches 73.49\%. 
For babble noise, degradation is even more severe, reaching 93.91\% and 137.05\% WER, respectively. We note that WER values above 100\% can occur due to a large number of insertion and substitution errors under extreme noise conditions.
In contrast, the visual-only model remains stable across all noise levels, achieving a constant WER of 47.80\%, as it does not depend on the audio signal.

Audio-visual fusion consistently improves robustness over audio-only models, particularly at low SNR levels. 
For example, under Gaussian noise at $-5$ dB, fusion reduces WER from 90.11\% to 80.26\% in the zero-shot setting, and from 73.49\% to 38.87\% in the fine-tuned setting. 
Similarly, under babble noise at $-5$ dB, fusion reduces WER from 93.91\% to 83.48\% (zero-shot) and from 137.05\% to 38.59\% (fine-tuned).
The gains from fusion decrease as SNR increases, as audio becomes more reliable. At higher SNR levels (10--15 dB), the gap between audio-only and fused models narrows, reflecting the diminishing contribution of the visual modality in clean conditions.

\vspace{0.3em}
\noindent{\bf Discussion.}
These results confirm that audio-visual fusion is particularly beneficial in challenging acoustic environments, where the audio signal is severely degraded. 
The visual modality provides a stable complementary signal that compensates for unreliable audio predictions.

Overall, our findings align with prior work~\cite{petridis2018end,martinez2020lipreading} showing that AVSR yields modest improvements under clean conditions but significantly enhances robustness under noise. 
This highlights the importance of evaluating multimodal systems in realistic noisy scenarios and demonstrates the effectiveness of confidence-aware fusion for robust speech recognition.

\subsection{Robustness to Domain Shift}

\noindent{\bf Setup.}
To evaluate robustness beyond the training distribution, we construct a set of OOD test categories from YouTube videos. 
These categories are designed to capture different types of domain shift: (1) \textit{Vlogs}: self-recorded videos with informal speaking style and varying camera setups; (2) \textit{Specific domains}: videos from specialized domains such as medical, engineering, religious, or linguistic content; (3) \textit{Noise}: videos with increased background noise and challenging acoustic conditions; (4) \textit{Black-and-white}: archival videos (e.g., 1950s–1970s) with degraded visual quality. Each category contains approximately 15 minutes of data, resulting in a total OOD test set of about 1 hour. We evaluate the best-performing model trained on the full 200 hours of pseudo-labeled data. 

\vspace{0.3em}
\noindent{\bf Results.}
Table~\ref{tab:ood_results} summarizes performance across OOD categories, together with out-of-vocabulary (OOV) statistics measured at both token and type level, where OOV token denotes the proportion of word occurrences not seen during training, and OOV type denotes the proportion of unique words absent from the training vocabulary.

\begin{table}[t]
\centering
\small
\caption{Performance on OOD categories. The model is trained on 200 hours of pseudo-labeled data. We also report out-of-vocabulary (OOV) statistics.}
\label{tab:ood_results}
\begin{tabular}{lcccc}
\toprule
\textbf{Category} & \textbf{\#Clips} & \textbf{WER (\%)} &  \textbf{OOV token (\%)} & \textbf{OOV type (\%)} \\
\midrule
Vlogs & 99 & 58.61 &  1.49 & 4.26 \\
Specific domains & 84 & 63.01 &  9.78 & 17.93 \\
Noise & 100 & 68.96 &  6.19 & 12.88 \\
Black-and-white & 92 & 87.97 &  5.24 & 10.96 \\
\midrule
\textbf{Global} & 375 & 68.46 &  5.08 & 14.75 \\
\bottomrule
\end{tabular}
\end{table}

\vspace{0.3em}
\noindent{\bf Discussion.} Performance varies substantially across OOD categories, reflecting different types of domain shift.
The model performs best on \textit{vlogs} (58.61\% WER), which are visually closest to the training data and exhibit the lowest OOV rates (1.49\% token-level), indicating good lexical coverage.

Performance degrades on \textit{specific domains} (63.01\% WER), which exhibit the highest OOV rates (9.78\% token-level, 17.93\% type-level). This suggests that domain-specific vocabulary plays a significant role in performance degradation, beyond visual or acoustic differences.

In the presence of \textit{noise} (68.96\% WER), performance further decreases, highlighting sensitivity to degraded acoustic conditions. This effect is compounded by a moderate increase in OOV rate, suggesting that both acoustic corruption and lexical mismatch contribute to errors.

The most challenging category is \textit{black-and-white} video (87.97\% WER), where degradation is primarily visual. Despite moderate OOV rates, performance drops sharply, indicating that the visual encoder is particularly sensitive to shifts in appearance, such as changes in resolution, contrast, and recording technology.

These results show that domain shift arises from multiple factors: vocabulary mismatch, acoustic degradation, and visual appearance changes. This highlights the importance of multimodal modeling and suggests that improving robustness requires addressing all three aspects in low-resource VSR.

\begin{table}[t]
\centering
\small
\setlength{\tabcolsep}{4pt}
\caption{Comparison on the LRRo isolated word benchmark. We report top-1 and top-5 accuracy on both Lab LRRo and Wild LRRo. Our models freeze the visual encoder and train only an MLP classifier on top of the extracted representations.}
\label{tab:lrro_results}
\begin{tabular}{llcccc}
\toprule
\textbf{Method} & \textbf{Preprocessing} & \textbf{Lab Acc@1} & \textbf{Lab Acc@5} & \textbf{Wild Acc@1} & \textbf{Wild Acc@5} \\
\midrule
Inception-V4~\cite{jitaru2020lrro} & -- & 71.0 & 92.0 & 33.0 & 62.0 \\
ResNet + BiGRU + LI~\cite{manescu2023end} & -- & -- & -- & 41.1 & 72.2 \\
ResNet + MS-TCN + LI~\cite{manescu2023end} & -- & -- & -- & 51.6 & 74.6 \\
\midrule
MultiVSR + MLP v1 & $96{\times}96$ resize & 90.6 & 98.5 & 64.5 & 87.6 \\
MultiVSR + MLP v2 & $64{\times}64$ center-middle & 91.4 & 99.0 & 68.6 & 89.3 \\
MultiVSR + MLP v3 & $64{\times}64$ center-bottom & \textbf{95.0} & \textbf{99.4} & \textbf{72.7} & \textbf{92.6} \\
MultiVSR + MLP v4 & $48{\times}48$ center-bottom & 93.1 & 99.3 & \textbf{72.7} & 90.9 \\
\bottomrule
\end{tabular}
\end{table}

\subsection{External Validation on LRRo}

\noindent{\bf Setup.}
To further evaluate whether the visual representations learned from VSRo-200 transfer beyond sentence-level transcription, we test them on the LRRo benchmark~\cite{jitaru2020lrro}, the first Romanian lip-reading dataset. LRRo contains two word-level subsets: \textit{Lab LRRo}, recorded in controlled conditions with 48 word classes, and \textit{Wild LRRo}, collected from Internet videos with 21 word classes. 

This evaluation differs from our main VSRo-200 setting in two important ways. First, LRRo is an isolated word classification benchmark, while VSRo-200 is designed for sentence-level VSR. Second, LRRo provides mouth crops of size $64 \times 64$, whereas our models are trained on larger face crops. To adapt our model to LRRo, we freeze the visual feature extractor and train a lightweight MLP classifier on top of the extracted representations.

\noindent{\bf Preprocessing variants.}
Because LRRo contains only mouth crops, we evaluate several resizing and positioning strategies before feeding the frames to the visual encoder:
(i) resizing the frames directly to $96 \times 96$;
(ii) keeping the $64 \times 64$ crop and padding it equally on all sides;
(iii) placing the $64 \times 64$ crop in the lower center of a $96 \times 96$ canvas. (iv) resizing the crop to $48 \times 48$ and placing it in the lower center of the canvas.
The last two variants are motivated by the attention patterns of the VTP encoder, which indicate that preserving the relative position of the mouth is important.

\noindent{\bf Results.}
Table~\ref{tab:lrro_results} shows that the representations learned from VSRo-200 transfer effectively to isolated word recognition on LRRo. On Lab LRRo, our best variants reach 95.0\% top-1 accuracy and 99.4\% top-5 accuracy, substantially outperforming the Inception-V4 baseline reported by Jitaru et al.~\cite{jitaru2020lrro}. On Wild LRRo, our best model reaches 72.7\% top-1 accuracy and 92.6\% top-5 accuracy, improving over the previously reported results of 51.6\% and 74.6\% obtained with ResNet-based models using cross-lingual domain adaptation and lateral inhibition~\cite{manescu2023end}.

The preprocessing strategy has a clear impact on performance. Directly resizing the $64 \times 64$ mouth crops to $96 \times 96$ already performs well, but positioning the mouth crop in the lower part of the input canvas yields better results. This suggests that preserving the spatial layout expected by the visual encoder is important for transferring representations across datasets with different crop definitions.

\noindent{\bf Discussion.}
These results provide external validation for VSRo-200. Although our dataset is designed for sentence-level VSR, the learned representations generalize well to a different Romanian benchmark, a different task formulation, and different visual preprocessing conditions. This supports the usefulness of VSRo-200 not only as a sentence-level dataset but also as a resource for learning transferable Romanian visual speech representations.

We further analyze robustness to demographic variation (male vs.\ female speakers) in the supplementary material, including per-group performance breakdowns.

\section{Limitations}

While VSRo-200 provides a large-scale benchmark for Romanian VSR, several limitations should be acknowledged.

\noindent\textbf{Domain and data diversity.}
The dataset is constructed from podcast videos, which exhibit relatively consistent visual conditions (e.g., frontal faces, stable lighting, and controlled camera setups). As a result, models trained on VSRo-200 may not generalize well to more unconstrained environments such as surveillance footage, spontaneous recordings, or extreme pose variations.

\noindent\textbf{Annotation quality.}
Although all data is annotated with pseudo-labels and a subset is manually transcribed, pseudo-labels inherently introduce noise. While our experiments show that scaling can compensate for this noise, performance may still degrade in scenarios with higher transcription errors or domain mismatch.

\noindent\textbf{Vocabulary coverage.}
Our OOD experiments indicate that performance is sensitive to out-of-vocabulary (OOV) words, particularly in domain-specific settings. This highlights a limitation in lexical coverage, which is especially pronounced in low-resource languages.

\noindent\textbf{Demographic balance.}
Despite efforts to include diverse speakers, the dataset may still exhibit imbalances in demographic attributes such as gender, age, or appearance. As shown in our supplementary analysis, models trained on skewed data may exhibit performance disparities across groups.

\noindent\textbf{Evaluation scope.}
Our experimental evaluation focuses on a single architecture (MultiVSR) and a limited number of training runs. While this allows controlled comparisons, the results may depend on the chosen model and training setup.

\noindent\textbf{Computational considerations.}
Training large-scale VSR models requires substantial computational resources, which may limit accessibility and reproducibility for some users.

\noindent\textbf{Privacy and ethical considerations.}
The dataset is constructed from publicly available videos featuring identifiable individuals. While we do not annotate sensitive attributes, models trained on this data may implicitly learn such information, raising potential concerns for misuse in surveillance or profiling applications.

\section{Conclusion}

We introduced \textit{VSRo-200}, the first large-scale dataset for Romanian VSR, comprising 200 hours of real-world podcast videos with both pseudo-labels and a subset of high-quality human annotations. The dataset enables controlled study of supervision regimes in a low-resource setting.
Through a comprehensive benchmark, we analyzed the impact of supervision quality, domain shift, and multimodal fusion. Our results show that pseudo-labels enable effective scaling, approaching the performance of human annotations at larger data sizes, while high-quality annotations remain important for robustness, particularly under domain shift. We further demonstrated that audio-visual fusion significantly improves performance in noisy conditions, and that domain generalization is influenced by multiple factors, including vocabulary mismatch, acoustic degradation, and visual variability.
Beyond sentence-level VSR, we showed that representations learned on VSRo-200 transfer effectively to the LRRo benchmark, substantially improving performance on isolated word recognition. 
Overall, VSRo-200 provides a new testbed for studying supervision, robustness, and multimodal learning in low-resource VSR. Future work includes improving robustness to visual and lexical domain shifts, as well as extending the dataset to more diverse and unconstrained scenarios.


\bibliographystyle{IEEEtran}
\bibliography{mybib}


\appendix

\section{Technical appendices and supplementary material}

\subsection{Ethical statement}
We share VSRo-200 under the Creative Commons Attribution Non-Commercial Share-Alike 4.0 (CC BY-NC-SA 4.0) license, aiming for open and responsible research. The dataset contains exclusively metadata derived from publicly available YouTube videos, including video IDs, textual transcription annotations, segment start and end timestamps, and speaker gender labels. No audio, visual, or personal content is stored or redistributed. We respect individual privacy rights, including the right to be forgotten. If any individual identifies themselves in the dataset and wishes to have their data removed, they can contact us and we will promptly address the request by removing the respective entries, in compliance with applicable data protection principles.

\subsection{Audio-Visual Fusion Details}
\begin{figure}[t]
    \centering
    \includegraphics[width=\linewidth]{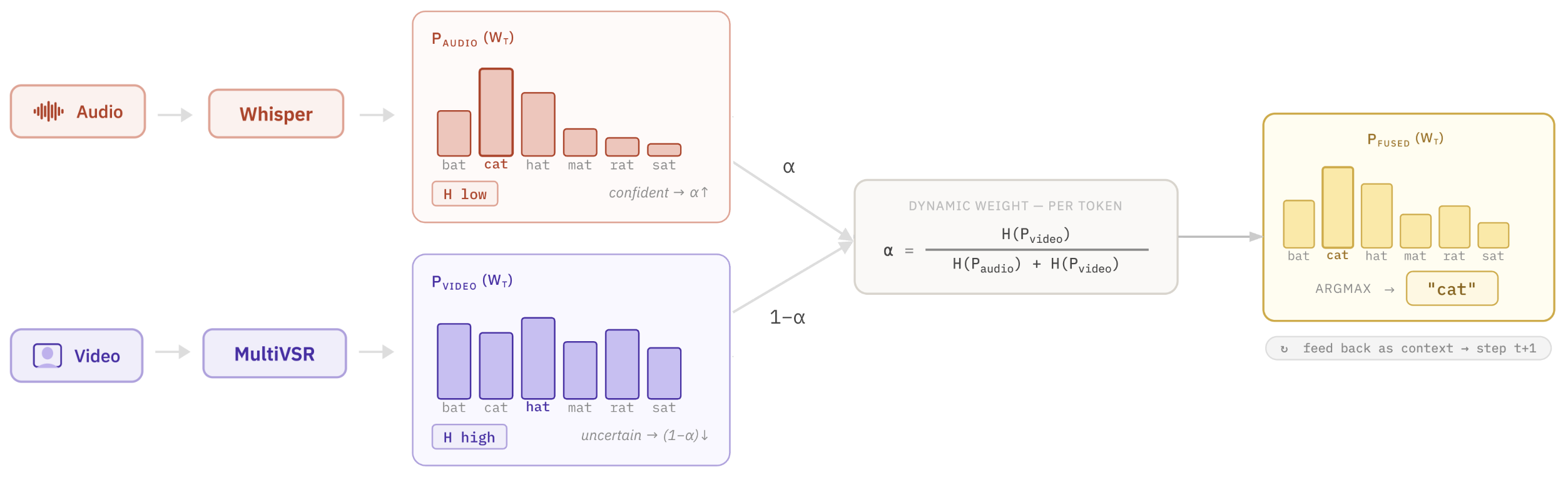}
    \caption{Overview of the proposed confidence-aware audio-visual fusion. At each decoding step, the audio and visual models produce probability distributions over the vocabulary. 
    We compute a dynamic weight $\alpha_t$ based on the entropy of each distribution, assigning higher importance to the more confident modality. 
    The fused distribution is then used to select the next token, which is fed back to both models.}
    \label{fig:avsr_fusion}
\end{figure}

Figure~\ref{fig:avsr_fusion} provides an overview of the proposed audio-visual fusion strategy described in Section~\ref{avsr}.

\begin{table*}[t!]
\centering
\small
\setlength{\tabcolsep}{4pt}
\caption{Robustness to demographic distribution. Models are trained on male-only, female-only, or mixed subsets (40 hours each). We report WER and CER globally, as well as separately for male (M) and female (F) speakers, under both seen and unseen settings.}
\label{tab:gender_bias}
\begin{tabular}{llcccccc}
\toprule
\textbf{Train} & \textbf{Test} & \textbf{WER} & \textbf{CER} & \textbf{WER (M)} & \textbf{CER (M)} & \textbf{WER (F)} & \textbf{CER (F)} \\
\midrule
\multirow{2}{*}{40h\_males}
& Unseen & 62.15 & 35.23 & 61.32 & 34.51 & 62.97 & 35.95 \\
& Seen   & 58.82 & 33.11 & 58.58 & 32.59 & 59.06 & 33.63 \\
\midrule
\multirow{2}{*}{40h\_females}
& Unseen & 59.33 & 33.44 & 59.17 & 32.87 & 59.49 & 34.02 \\
& Seen   & 59.10 & 33.30 & 67.26 & 38.67 & 51.20 & 27.99 \\
\midrule
\multirow{2}{*}{40h\_mix}
& Unseen & 59.52 & 33.74 & 59.19 & 33.26 & 59.85 & 34.22 \\
& Seen   & 56.29 & 31.22 & 60.56 & 33.54 & 52.15 & 28.93 \\
\bottomrule
\end{tabular}
\end{table*}

\subsection{Robustness to Demographic Distribution}

To complement the main experiments, we analyze robustness with respect to demographic variation, focusing on potential performance differences across speaker gender.

\noindent\textbf{Setup.}
To analyze robustness with respect to demographic variation, we evaluate models trained on three different subsets of the data: (i) male-only speakers (40h\_males), (ii) female-only speakers (40h\_females), and (iii) a mixed subset (40h\_mix). Each model is evaluated on both \textit{seen} and \textit{unseen} speakers. In addition to global WER and CER, we report performance separately for male (M) and female (F) speakers in the test set.

\noindent\textbf{Results.}
Table~\ref{tab:gender_bias} summarizes the results. We observe consistent differences in performance depending on both the training distribution and the demographic composition of the test data.

First, models trained on male-only data (40h\_males) exhibit slightly better performance on male speakers compared to female speakers (e.g., 61.32\% vs.\ 62.97\% WER on unseen speakers), indicating a mild bias toward the dominant training distribution. A similar trend is observed for female-only training, although the effect is less consistent across settings.

Second, the largest discrepancy appears in the \textit{seen} speaker setting for the model trained on female-only data, where performance differs substantially between male and female speakers (67.26\% vs.\ 51.20\% WER). This suggests that models trained on a single demographic group may overfit to specific visual characteristics and fail to generalize uniformly across groups.

Third, training on mixed data (40h\_mix) leads to the most balanced performance across genders. While not always achieving the lowest WER, the mixed setting reduces the gap between male and female performance (e.g., 60.56\% vs.\ 52.15\% WER on seen speakers), indicating improved robustness.

\noindent\textbf{Discussion.}
These results highlight the importance of demographic diversity in training data for VSR. Models trained on homogeneous data tend to specialize in the dominant visual patterns, which may limit their ability to generalize across demographic groups.

A possible explanation is that certain visual attributes (e.g., facial hair, lip visibility, or articulation differences) introduce systematic variations in the visual signal. For instance, the presence of facial hair in male speakers may partially occlude the lip region, leading the model to rely on features that do not transfer well to female speakers.

Overall, these findings suggest that balanced and diverse training data is essential for building robust and fair VSR systems. Future work should further investigate normalization strategies or augmentation techniques to mitigate demographic biases.



\end{document}